\documentclass[conference]{IEEEtran}
\IEEEoverridecommandlockouts
\usepackage{cite}
\usepackage{amsmath,amssymb,amsfonts}
\usepackage{algorithmic}
\usepackage{graphicx}
\usepackage{textcomp}
\usepackage{xcolor}

\usepackage{mathrsfs,amssymb,amsmath}
\usepackage{cite}

\usepackage{mathtools} 
\usepackage{todonotes}
\usepackage{tabularx} 
\usepackage{booktabs}
\usepackage[linesnumbered,ruled,vlined]{algorithm2e}
\SetKwInOut{Parameter}{Parameter}

\SetKwInput{KwInput}{Input}                
\SetKwInput{KwOutput}{Output}              
\usepackage{hyperref}

\usepackage{tikz, pgfplots}
\usetikzlibrary{automata,positioning,arrows,arrows.meta,shapes,decorations.pathreplacing,fit,calc}
\usetikzlibrary{automata,positioning,arrows,shapes,decorations.pathreplacing,fit,calc}
\usetikzlibrary{angles,quotes}
\usetikzlibrary{decorations.pathreplacing,calc}  
\usetikzlibrary{matrix,positioning, arrows.meta}
\tikzset{signal/.style={draw,thick,minimum width=2cm,minimum height=1cm}}
\tikzset{box/.style={draw,thick,minimum width=1cm,minimum height=1cm}}
\tikzset{arrow/.style={draw,thick,->}}
\usepackage{environ}
\usepackage{placeins}
\usepackage{makecell}

\usepackage{tikz}
\usetikzlibrary{positioning}
\usetikzlibrary{decorations.pathreplacing}
\usetikzlibrary{shapes,snakes}
\usetikzlibrary{calc}
\usetikzlibrary{backgrounds}
\usepackage{adjustbox}
\usepackage{bm} 

\def\BibTeX{{\rm B\kern-.05em{\sc i\kern-.025em b}\kern-.08em
    T\kern-.1667em\lower.7ex\hbox{E}\kern-.125emX}}

\newcommand\copyrighttext{%
	\footnotesize \copyright~2024 IEEE. Personal use of this material is permitted. Permission from IEEE must be obtained for all other uses, in any current or future media, including reprinting/republishing this material for advertising or promotional purposes,creating new collective works, for resale or redistribution to servers or lists, or reuse of any copyrighted component of this work in other works.}
\newcommand\copyrightnotice{%
	\begin{tikzpicture}[remember picture,overlay]
		\node[anchor=south,yshift=10pt] at (current page.south) {\fbox{\parbox{\dimexpr\textwidth-\fboxsep-\fboxrule\relax}{\copyrighttext}}};
	\end{tikzpicture}%
}

\begin{document}

\title{Spectral Wavelet Dropout: Regularization in the Wavelet Domain}

\author{\IEEEauthorblockN{Rinor Cakaj}
	\IEEEauthorblockA{\textit{Signal Processing} \\
		\textit{Robert Bosch GmbH \& University of Stuttgart}\\
		71229 Leonberg, Germany \\
		Rinor.Cakaj@de.bosch.com}
	\and
	\IEEEauthorblockN{Jens Mehnert}
	\IEEEauthorblockA{\textit{Signal Processing} \\
		\textit{Robert Bosch GmbH}\\
		71229 Leonberg, Germany \\
		JensEricMarkus.Mehnert@de.bosch.com}
	\and
	\IEEEauthorblockN{Bin Yang}
	\IEEEauthorblockA{\textit{ISS} \\
		\textit{University of Stuttgart}\\
		70550 Stuttgart, Germany \\
		bin.yang@iss.uni-stuttgart.de}
}

\maketitle

\begin{abstract}
	Regularization techniques help prevent overfitting and therefore improve the ability of convolutional neural networks (CNNs) to generalize. One reason for overfitting is the complex co-adaptations among different parts of the network, which make the CNN dependent on their joint response rather than encouraging each part to learn a useful feature representation independently. Frequency domain manipulation is a powerful strategy for modifying data that has temporal and spatial coherence by utilizing frequency decomposition. This work introduces Spectral Wavelet Dropout (SWD), a novel regularization method that includes two variants: 1D-SWD and 2D-SWD. These variants improve CNN generalization by randomly dropping detailed frequency bands in the discrete wavelet decomposition of feature maps. Our approach distinguishes itself from the pre-existing Spectral ``Fourier'' Dropout (2D-SFD), which eliminates coefficients in the Fourier domain. Notably, SWD requires only a single hyperparameter, unlike the two required by SFD. We also extend the literature by implementing a one-dimensional version of Spectral ``Fourier'' Dropout (1D-SFD), setting the stage for a comprehensive comparison. Our evaluation shows that both 1D and 2D SWD variants have competitive performance on CIFAR-10/100 benchmarks relative to both 1D-SFD and 2D-SFD. Specifically, 1D-SWD has a significantly lower computational complexity compared to 1D/2D-SFD. In the Pascal VOC Object Detection benchmark, SWD variants surpass 1D-SFD and 2D-SFD in performance and demonstrate lower computational complexity during training.
\end{abstract}
\copyrightnotice

\section{Introduction} \label{Introduction}

Over recent years, convolutional neural networks (CNNs) have demonstrated significant advancements in a variety of computer vision applications, such as image recognition \cite{2016_He_CONF} or object detection \cite{2015_Ren_CONF}. However, CNNs are prone to overfitting, especially when the available 
training data is limited or when the model's capacity is excessively large~\cite{2013_Wan_CONF}. Regularization techniques are necessary 
to ensure that models can generalize effectively to new, unseen data~\cite{2016_Goodfellow_BOOK}.

Various regularization methods have been developed to address the overfitting problem in CNNs. These include weight penalties such as $L_1$ regularization~\cite{1996_Tibshirani} 
and weight decay~\cite{1991_Krogh_CONF}, in addition to techniques like data augmentation~\cite{2019_Shorten}, 
and ensemble learning methods~\cite{1999_Opitz}. Among these, dropout mechanisms, notably standard dropout~\cite{2014_Srivastava} and dropconnect~\cite{2013_Wan_CONF}, 
play an important role by randomly dropping neurons, weights, or feature map entries during training to prevent co-adaptation of units within the network. 
However, the effectiveness of these methods is often compromised in convolutional layers where features have spatial correlation, leading to suboptimal regularization 
and the continuation of overfitting problems~\cite{2018_Ghiasi_CONF}.

The limitations within standard dropout approaches have led to the development of more advanced variants like dropblock~\cite{2018_Ghiasi_CONF} and dropout2d~\cite{2015_Tompson_CONF}. These methods introduce structured forms of dropout, designed to improve regularization in CNNs by strategically dropping feature map segments or channels. This addresses the spatial correlation in the feature maps and therefore increases the regularization effect.

\begin{figure*}[tb]
	\includegraphics[width=\linewidth]{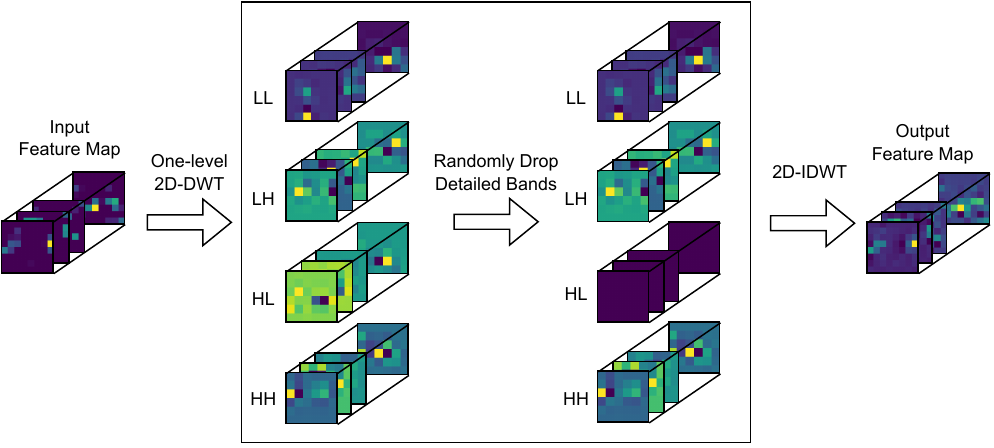}
\caption{Illustration of the 2D Spectral Wavelet Dropout process applied to an input feature map. A one-level 2D wavelet decomposition divides the map into four sub-bands: LL (low-frequency), LH (horizontal high-frequency), HL (vertical high-frequency), and HH (diagonal high-frequency), presented sequentially from top to bottom in the figure. The high-frequency bands are randomly dropped (HL is shown as dropped) according to a dropout probability \( p \), while the LL band is consistently maintained. The feature map is then transformed back into the spatial domain using an inverse 2D wavelet transform (2D-IDWT).}
	
	\label{fig:SWD_Layer}
\end{figure*}

The investigation of alternative regularization strategies has led to the use of frequency domain manipulation. Frequency decomposition provides a natural framework for the manipulation of data with
spatial coherence \cite{2015_Rippel_CONF}. Khan et al. \cite{2019_Khan} introduced Spectral ``Fourier'' Dropout (2D-SFD), a strategy designed to reduce overfitting by selectively pruning and dropping Fourier domain coefficients within feature maps. The process begins with the application of a two-dimensional discrete cosine transform (2D-DCT) to transform each channel in feature maps into the frequency domain. It then prunes Fourier coefficients that are considered weak or ``noisy'', i.e.\ those that fall below the $\eta$ quantile. Subsequently, a dropout mask is applied to the remaining coefficients with a dropout probability $p$. Finally, the modified feature map is transformed back into the spatial domain using an inverse 2D-DCT. The effectiveness of 2D-SFD lies in its ability to make CNNs insensitive to weak and ``noisy'' spectral components. This is achieved by randomly preserving only the important Fourier coefficients for signal reconstruction. Moreover, it effectively prevents co-adaptation among feature detectors by diversifying the network's reliance across various frequency components~\cite{2019_Khan}.

The discrete wavelet transform (DWT) provides an alternative way for regularization, offering advantages in handling multi-resolution data. 
Unlike the Fourier transform, which decomposes signals into sinusoidal components without localization, the DWT divides data into frequency bands at multiple resolutions, 
capturing both global and localized features~\cite{2016_Williams_CONF}. This capability motivates the investigation of Spectral Wavelet Dropout (SWD), 
a novel method proposed in this work for improving CNN generalization by randomly dropping wavelet bands within the discrete wavelet decomposition of feature maps.

We introduce two variants of Spectral Wavelet Dropout: the one-dimensional (1D-SWD) and two-dimensional (2D-SWD) versions. For 1D-SWD, consider a feature map $X \in \mathbb{R}^{C \times H \times W}$. The initial step involves flattening the feature map across each channel, resulting in $\hat{X} \in \mathbb{R}^{C \times H \cdot W}$. Then, 1D-SWD performs a three-level one-dimensional discrete wavelet decomposition (1D-DWT) of $\hat{X}$. This decomposition produces approximation and detailed frequency coefficients across three levels, also referred to as frequency bands. In the next phase, 1D-SWD randomly drops detailed wavelet bands based on a dropout probability $p$, effectively introducing regularization. The final step involves transforming the feature map back into the spatial domain through an inverse 1D-DWT and reshaping it for input into the subsequent network layer. An illustration of 1D-SWD is shown in Figure \ref{fig:SWD_Layer_app} in the Appendix.

The 2D-SWD method applies a one-level two-dimensional wavelet decomposition to each channel of the feature map $X$. This decomposition process produces four distinct sub-bands: LL (approximation), LH (horizontal details), HL (vertical details), and HH (diagonal details). Following the 1D-SWD method, 2D-SWD randomly drops the LH, HL, and HH detailed frequency bands, based on a dropout probability $p$, while preserving the LL approximation band. The modified feature map is then transformed back into the spatial domain through an inverse 2D wavelet transform (2D-IDWT) before being input into the subsequent network layer. Both 1D-SWD and 2D-SWD are characterized by a single hyperparameter, $p$, which considerably simplifies the hyperparameter tuning process. An illustration of 2D-SWD is shown in Figure \ref{fig:SWD_Layer}.

Jo and Bengio~\cite{2017_Jo} showed that CNNs have a bias towards higher frequency components. Similarly, Geirhos et al.~\cite{2019_Geirhos_CONF} presented a texture bias in CNNs trained on ImageNet \cite{2015_Russakovsky}. Building on these findings, our method differs from the Spectral ``Fourier'' Dropout \cite{2019_Khan} approach by strategically targeting the frequency domain. Specifically, SWD randomly drops detailed frequency bands. This approach is based on the understanding that the approximation band of the wavelet transformation typically provides a smoothed representation of the original signal, while the detailed bands capture high frequency variation and noise. By randomly dropping the detailed bands, SWD encourages the CNN to diversify its reliance away from single detailed frequency bands within the feature maps.

We extend the literature by implementing a one-dimensional version of Spectral ``Fourier'' Dropout (1D-SFD), to present a fair comparison of the methods. As a one-dimensional analogue to 2D-SFD, 1D-SFD begins with the flattening of the feature map across each channel. This is followed by the application of a one-dimensional discrete cosine transform (1D-DCT). Subsequently, a specified proportion of activations, denoted by $\eta$, is pruned, and a dropout mask is applied to the remaining Fourier coefficients using a dropout probability $p$. Finally, the modified feature map is transformed back to the spatial domain using the inverse 1D-DCT. It is noteworthy that 1D-SFD uses two hyperparameters.

The motivation behind using 1D transformations lies in their computational efficiency. Moreover, they offer specific pattern recognition capabilities for images with informative horizontal features. While 2D transformations provide a more detailed analysis by capturing both vertical and horizontal patterns, our experiments demonstrate that 1D approaches can achieve comparable regularization effects with lower computational demands.

Our evaluation shows that both 1D and 2D-SWD variants have competitive performance on CIFAR-10/100 benchmarks relative to both 1D-SFD and 2D-SFD. Specifically, 1D-SWD is highlighted for its competitive results, considering its significantly lower computational complexity during training compared to 1D/2D-SFD. On ImageNet, 1D-SWD maintains competitive performance considering the reduced computational demand during training. In the Pascal VOC Object Detection benchmark, SWD variants exceed 1D-SFD and 2D-SFD in terms of performance while having a lower training overhead. Specifically, 2D-SWD improves the mean Average Precision (mAP) by 0.56 percentage points (a 0.72\% increase) compared to the baseline. A key benefit of the SWD method is its simplified setup, which only requires one hyperparameter, unlike the SFD method that needs two. 

\section{Related Work} \label{Related-Work}

\subsection{Regularization}

Regularization is one of the key elements of deep learning
\cite{2016_Goodfellow_BOOK}, allowing the model to generalize well to unseen
data even when trained on a finite training set or with an imperfect
optimization procedure \cite{2017_Kukacka}. There are several techniques to
regularize CNNs which can be categorized into groups. Data augmentation
methods like cropping, flipping, adjusting brightness or sharpness
\cite{2019_Shorten} and cutout \cite{2017_DeVries}
transform the training dataset to avoid overfitting.
Regularization techniques like dropout \cite{2014_Srivastava},
dropblock \cite{2018_Ghiasi_CONF} or dropconnect
\cite{2013_Wan_CONF} drop neurons or weights from the CNN during training to
prevent units from co-adapting too much \cite{2014_Srivastava}. Khan et al.\
\cite{2019_Khan} extend the idea of dropping parts of CNNs to the
spectral domain. Spectral ``Fourier'' Dropout avoids overfitting by pruning weak and ``noisy'' Fourier domain coefficients of the
feature maps and randomly dropping a percentage of the remaining Fourier domain coefficients.

Furthermore, CNNs can be regularized using penalty terms in the loss function. Weight decay \cite{1991_Krogh_CONF} encourages the weights of the CNN to be small in magnitude. The $L_1$-regularization \cite{1996_Tibshirani} encourages the weights of non-relevant features to be zero. 

\subsection{Wavelet Transforms in CNNs}

Wavelet transformations are used in image processing \cite{2022_Finder}, for example the JPEG2000 format \cite{2002_Taubman_BOOK} uses wavelets to represent
images as highly sparse feature maps.  Recently, wavelet transforms have been used in deep learning due to their multi-resolution analysis properties. 

\subsubsection{Wavelet Neural Networks}

In a limited form, CNNs can be seen as a form of multi-resolution analysis.
However, conventional CNNs are missing a large part of spectral information present in feature maps \cite{2018_Fujieda}. Therefore, different approaches supplement the missing part
of the spectral information via additional components in the entire architecture
using the DWT. Fujieda et al.\ \cite{2018_Fujieda} present wavelet CNNs and
demonstrate benefits in texture classification. Liu et al.\
\cite{2019_Liu} integrates a wavelet transform into the CNN architecture to reduce the resolution of feature maps, while at the same time increasing the receptive field.  

\subsubsection{Pooling}

Depending on the data, max pooling can remove and average pooling can dilute
important details from an image. Wavelet Pooling \cite{2018_Williams_CONF}
overcomes these issues by decomposing the feature maps into a second-level
decomposition and removing the first-level sub-bands to reduce the feature
dimensions. Adaptive wavelet pooling \cite{2021_Wolter_CONF} extends this
approach by using adaptive- and scaled-wavelets in the pooling process. 

\subsubsection{Compression}

CNNs are successfully used in a number of applications. Despite this, their storage
requirements have largely prevented their use on mobile devices.
Wolter et al.\ \cite{2020_Wolter_CONF} showed how the fast wavelet transform can
be used to compress linear layers in neural networks using learnable wavelets.

\section{Method} \label{Method}

One of the main advantages of the wavelet transform over the Fourier transform is its superior spatial/temporal resolution. This property allows the wavelet transform to capture information about both frequency and location simultaneously, providing an optimal structure for manipulating data with spatial coherence.

Inspired by the concept of Spectral ``Fourier'' Dropout presented by Khan et al.~\cite{2019_Khan}, we introduce Spectral Wavelet Dropout, which includes two variants: 1D-SWD and 2D-SWD. These variants improve CNN generalization by randomly dropping detailed frequency bands in the discrete wavelet decomposition of feature maps. To effectively integrate SWD into CNNs, we first look at the basics of one- and two-dimensional discrete wavelet transforms.

\subsection{Discrete Wavelet Transform} \label{DWT_Theory}


The discrete wavelet transform is a method for analyzing signals and images using discretely sampled wavelets. The DWT can be implemented with various wavelet families, such as Haar wavelets~\cite{1910_Haar} and Daubechies wavelets~\cite{1992_Daubechies_BOOK}.

Given a vector $x \in \mathbb{R}^d$ of even length, the one-level one-dimensional DWT (1D-DWT) is computed by processing $x$ through corresponding low-pass and high-pass filters, denoted as $g$ and $h$. These filters, which are quadrature mirror filters~\cite{2004_Mohlenkamp}, effectively split the frequency spectrum of the signal, allowing every other sample to be removed according to Nyquist's theorem~\cite{1949_Shannon}. The computation is formalized by the following equations:
\begin{align} \label{dwt_formulas}
	y_{\text{low}}[n] &= (g * x)[n] = \sum_{k=0}^{N-1} g[k] \cdot x[2n-k], \\
	y_{\text{high}}[n] &= (h * x)[n] = \sum_{k=0}^{N-1} h[k] \cdot x[2n-k],
\end{align}
where $N$ is the filter length and $n$ denotes the output terms, with $n \in \{1, \dots, d/2\}$. The output $y_{\text{low}}$ represents the approximation coefficients from the low-pass filter and $y_{\text{high}}$ the detailed coefficients from the high-pass filter.

1D-DWT requires signals with lengths that are powers of two. To address border distortions and signal lengths that do not meet this condition, the signal is typically extended. One common method is zero-padding, where zeros are appended to the signal to meet the required length for transformation.

\begin{figure}
	\centering
	\resizebox{\columnwidth}{!}{%
		\begin{tikzpicture}[node distance = 2cm, auto, scale=0.5]
			\node[rectangle] (input) {$x[n]$};
			
			\node[box, above right = 1cm and 1 cm of input] (g0) {$g[n]$};
			\node[box, right= 1cm of input] (h0) {$h[n]$};
			\node[circle, draw, right = 1cm of g0] (dsg0) {$\downarrow 2$};
			\node[circle, draw, right = 1cm of h0] (dsh0) {$\downarrow 2$};
			\node[rectangle, right = 1cm of dsh0, align=left] (lv1) {Level 1 \\ coefficients};
			
			\path [arrow] (input) -| ($(input)+(1.8,0)$) |- (g0);
			\path [arrow] (input) -| ($(input)+(1.8,0)$) |- (h0);
			\path [arrow] (g0) |- (dsg0);
			\path [arrow] (h0) |- (dsh0);
			\path [arrow] (dsh0) -- (lv1);
			
			\node[box, right = 1cm of dsg0] (h1) {$h[n]$};
			\node[box, above = 1cm of h1] (g1) {$g[n]$};
			\node[circle, draw, right = 1cm of g1] (dsg1) {$\downarrow 2$};
			\node[circle, draw, right = 1cm of h1] (dsh1) {$\downarrow 2$};
			\node[rectangle, right = 1cm of dsh1, align=left] (lv2) {Level 2 \\ coefficients};
			
			\path [arrow] (dsg0) -| ($(dsg0)+(1.8,0)$) |- (g1);
			\path [arrow] (dsg0) -| ($(dsg0)+(1.8,0)$) |- (h1);
			\path [arrow] (g1) |- (dsg1);
			\path [arrow] (h1) |- (dsh1);
			\path [arrow] (dsh1) -- (lv2);
			
			\node[box, right = 1cm of dsg1] (h2) {$h[n]$};
			\node[box, above = 1cm of h2] (g2) {$g[n]$};
			\node[circle, draw, right = 1cm of g2] (dsg2) {$\downarrow 2$};
			\node[circle, draw, right = 1cm of h2] (dsh2) {$\downarrow 2$};
			\node[rectangle, right = 1cm of dsg2] (inlv3) {};
			\node[rectangle, right = 1cm of dsh2] (in1lv3) {};
			\node[rectangle, right = 1cm of dsh2, align=left] (in2lv3) {Level 3 \\ coefficients};
			\node[rectangle, right = 1cm of dsg2, align=left] (in2lv4) {Approximation \\ coefficients};
			
			\path [arrow] (dsg1) -| ($(dsg1)+(1.8,0)$) |- (g2);
			\path [arrow] (dsg1) -| ($(dsg1)+(1.8,0)$) |- (h2);
			\path [arrow] (g2) |- (dsg2);
			\path [arrow] (h2) |- (dsh2);
			\path [arrow] (dsg2) -- (inlv3);
			\path [arrow] (dsh2) -- (in1lv3);
			
		\end{tikzpicture}
	}
	\caption{Detailed schematic of a three-level wavelet filter bank used in our proposed 1D-SWD method. The filter bank decomposes an input signal \( x[n] \) into hierarchical sets of approximation and detailed coefficients. This allows analysis of the signal across multiple resolutions. This structured arrangement, using sequential high-pass \( h[n] \) and low-pass \( g[n] \) filters with subsequent downsampling, enables SWD to prevent overfitting. By randomly dropping high-frequency details, it promotes model generalization and prevents overfitting.}
	\label{fig:filter_bank}
\end{figure}
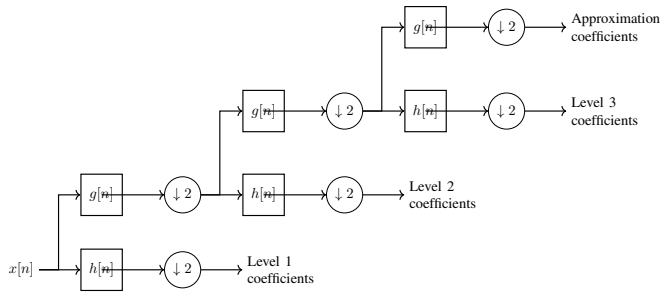

The 1D-DWT's decomposition can be iteratively applied to the approximation coefficients to refine frequency resolution. Each iteration applies high- and low-pass filters to the signal followed by downsampling. This process can be visualized in a binary tree structure, often referred to as a filter bank (as shown in Figure~\ref{fig:filter_bank}). The downsampling operation is symbolized by $\downarrow 2$. The approximation coefficients are a smoothed version of the original signal, whereas the detailed coefficients at levels 1, 2, and 3 capture increasingly fine-grained high-frequency content and noise.

The two-dimensional discrete wavelet transform (2D-DWT) is used to analyze signals in two-dimensional spaces, such as images. This transformation is applied sequentially along two orthogonal dimensions, commonly rows and columns. Firstly, a one-level one-dimensional DWT is applied to each row of the image matrix, decomposing it into low and high-frequency components. Then, this decomposition is applied to each column of the resulting sub-images, leading to four distinct sub-bands for each level of decomposition:

\begin{itemize}
	\item $LL$ (Low-Low) sub-band: This contains the approximation coefficients, resulting from applying the low-pass filter across both dimensions. It shows the primary features of the image, serving as a blurred, downsampled version of the original.
	\item $LH$ (Low-High) sub-band: This contains the vertical details of the image, derived from applying the low-pass filter horizontally and the high-pass filter vertically.
	\item $HL$ (High-Low) sub-band: This contains horizontal details, derived by applying the high-pass filter horizontally and the low-pass filter vertically.
	\item $HH$ (High-High) sub-band: This sub-band contains diagonal details, derived by applying the high-pass filter across both dimensions.
\end{itemize}

The DWT operates through matrix multiplication, characterizing it as a linear transformation. This linearity is essential because, during backpropagation, the gradient computation with respect to the DWT is directly determined by the transform matrix itself. 

\subsection{Spectral Wavelet Dropout} \label{SWD}

Spectral Wavelet Dropout is introduced in two variants, 1D-SWD and 2D-SWD.
Given a batch of feature maps $X \in \mathbb{R}^{B \times C \times H \times W}$, 1D-SWD begins by flattening $X$ along the spatial dimensions, resulting in $\hat{X} \in \mathbb{R}^{B \times C \times H \cdot W}$. A three-level 1D-DWT is performed using the Daubechies-3 (db3) wavelet~\cite{1992_Daubechies_BOOK}, producing approximation and detailed frequency bands. 

Subsequently, a mask $m \in \{0,1\}^{3}$ is generated, where each element $m_i$ is drawn from a Bernoulli distribution with a probability of $(1-p)$ for being 1. This mask is used to selectively drop detailed frequency bands. The approximation coefficients are preserved to maintain the primary signal structure. To compensate for the effect of dropout on signal energy, the feature maps are scaled by a factor of $(1-p)^{-1}$.

The process for 2D-SWD is analogous. A one-level 2D-DWT is applied to each channel, resulting in LL, LH, HL, and HH sub-bands. The mask is then applied to the detailed sub-bands.

Both processes end with the inverse DWT, 1D-IDWT for 1D-SWD and 2D-IDWT for 2D-SWD, followed by reshaping to the original dimensions in the 1D-DWT variant. Algorithm \ref{algorith_SWD} shows the steps for the 1D-SWD. The complete procedure for the 2D-SWD is explained in Section \ref{algorithm_2d_dwt} the Appendix.

The principle of SWD is based on the ability of the wavelet transformation to decompose signals into components of different spatial-frequency localization. By selectively dropping the detailed frequency components, SWD encourages the CNN to not overly rely on any specific set of frequencies, thereby preventing co-adaptation of feature detectors. This promotes a learning of features across all spatial frequencies present in the data.

\begin{algorithm}[!ht] 
	\DontPrintSemicolon
	\Parameter{\text{dropout probability = }$p \in (0,1)$}
	\KwInput{$X \in \mathbb{R}^{B \times C \times H \times W}$}
	\KwOutput{$\hat{X} \in \mathbb{R}^{B \times C \times H \times W}$}
	
	\tcp{Save size of $X$ and flatten the last two dimensions of $X$}
	size = $X$.size()
	
	$\hat{X} = \text{flatten}(X, \text{start\_dim}=2)$
	
	\tcp{Compute the 3-level 1D-DWT for each channel using the daubechie-3 wavelet}
	$AP, L1, L2, L3 = \text{1D-DWT}(\hat{X}, J=3, \text{wave='db3'})$
	
	\tcp{Create mask for dropout}
	
	$\text{dropout\_mask} = \text{Bernoulli}(3, \text{prob}=1-p)$
	
	\tcp{Wavelet Dropout}
	
	$L1 = L1 \cdot \text{dropout\_mask}[0]/(1-p)$
	
	$L2 = L2 \cdot \text{dropout\_mask}[1]/(1-p)$
	
	$L3 = L3 \cdot \text{dropout\_mask}[2]/(1-p)$
	
	\tcp{Compute inverse Wavelet transformation}
	
	$\hat{X} =  \text{1D-IDWT}([AP, L1, L2, L3], \text{wave='db3'})$
	
	\tcp{Reshape the feature map}
	
	$\hat{X} = \text{reshape}(\hat{X}, \text{size})$
	
	\caption{Spectral Wavelet Dropout - 1D-SWD} \label{algorith_SWD}
\end{algorithm}

\section{Experiments} \label{Experiments}

We evaluated the Spectral Wavelet Dropout method on supervised image recognition on CIFAR-10/100 \cite{2009_Krizhevsky} and ImageNet \cite{2015_Russakovsky} datasets, and object detection on the Pascal VOC \cite{2010_Everingham} dataset.

The implementation of 1D/2D-SFD \cite{2019_Khan} was supported by the PyTorch DCT library.\footnote{\url{https://github.com/zh217/torch-dct}} For 1D/2D-SWD, we utilized the wavelet transform provided by Cotter \cite{2019_Cotter_DISSERTATION}.

A notable difference in our approach compared to SFD is that 1D/2D-SWD requires only a single hyperparameter - the dropout rate $p$. We performed a hyperparameter search for $p$ within the set $p=\{0.1, 0.2, 0.3, 0.4, 0.5\}$. In contrast, for 1D/2D-SFD, a grid search was performed over both the dropout rate $p=\{0.1, 0.2, 0.3, 0.4, 0.5\}$ and the pruning rate $\eta=\{0, 0.1, 0.2, 0.3,
0.4\}$. Including $\eta=0$ allowed us to evaluate the efficacy of 1D/2D-SFD under a single hyperparameter setting, although this did not consistently yield optimal results. Detailed outcomes of these hyperparameter searches and further implementation details are provided in the Appendix \ref{hyperparameters_paper}.

Training overheads are reported relative to the baseline model to offer a clear comparison of the additional computational resources required. Further details on the methodology for runtime measurement are provided in Section \ref{comp_complexity}.

\subsection{Image Classification on CIFAR-10/100}

We analyzed the impact of both 1D-SWD and 2D-SWD on different architectures, including ResNet18 (R18), ResNet34 (R34), ResNet50 (R50) \cite{2016_He_CONF}, and VGG16-BN (V16) \cite{2015_Simonyan_CONF}, focusing on accuracy improvements and training overhead. The results are shown in Tables \ref{tab:cifar10-100-1D} and \ref{tab:cifar10-100-2D}.

Residual Networks (ResNets) \cite{2016_He_CONF} are structured into four stages consisting of basic and bottleneck blocks, starting with an initial convolutional layer and ending with a terminal linear layer. Analogous to other dropout variants like dropout2d \cite{2015_Tompson_CONF}, dropblock \cite{2018_Ghiasi_CONF}, and Spectral ``Fourier'' Dropout \cite{2019_Khan}, we integrate 1D/2D-SWD within the deeper layers of the CNNs to improve regularization. Implementation details are provided in the Appendix, Section \ref{cifar_implementation_details}.

\subsubsection{Results}

The results of our experiments are shown in Table \ref{tab:cifar10-100-1D} and \ref{tab:cifar10-100-2D}. The use of 1D-SWD resulted in accuracy improvements for all models compared to the baseline configurations. Specifically, the accuracy of ResNet50 on CIFAR-100 was improved by 1.28\%pt. Additionally, integrating 2D-SWD on ResNet50 for CIFAR-100 improved performance by 1.8\%pt.

Our analysis shows that on the CIFAR-10 dataset, both 1D and 2D Spectral Wavelet Dropout methods demonstrate performance that is competitive with, and sometimes even surpasses, that of 1D and 2D Spectral Fourier Dropout. For example, ResNet34 with 1D-SWD achieved an accuracy of $94.41\%\pm0.10\%$, outperforming 1D and 2D-SFD. 

While 2D Spectral Wavelet Dropout offers notable accuracy improvements, it also has a higher computational overhead compared to its 1D counterpart. Moreover, compared to the overheads associated with SFD methods, 1D-SWD demonstrates the lowest training overhead. This highlights the computational efficiency of 1D-SWD, making it a more practical choice for real-world applications where computational resources are constrained.

The regularization capability of 1D-SWD, demonstrated to be on par with or superior to 1D/2D-SFD. The combination of its computational efficiency and simplicity makes it an optimal choice for tasks where both performance and computational efficiency are valued. 

\begin{table}[tb]
	\caption{1D Spectral Dropout Methods on CIFAR-10/100: Accuracy and Training Time Multiplier (TTM). \label{tab:cifar10-100-1D}}
	\begin{tabular*}{\linewidth}{@{\extracolsep{\fill}}lccc} 
		\toprule
		Method & CIFAR-10 Acc. & CIFAR-100 Acc. & TTM \tabularnewline
		\midrule
		R18  & $94.04\% \pm 0.08\%$ & $76.47\% \pm 0.20\%$ & 1.00x\tabularnewline
		R18 + 1D-SFD & $94.19\% \pm 0.31\%$ & $\mathbf{77.41\% \pm 0.13\%}$ & 1.15x \tabularnewline
		R18 + 1D-SWD & $\mathbf{94.35\% \pm 0.13\%}$ & $77.07\% \pm 0.18\%$ & 1.12x \tabularnewline
		\midrule
		R34  & $93.69\% \pm 0.30\%$  &  $77.07\% \pm 0.45\%$ & 1.00x \tabularnewline
		R34 + 1D-SFD & $93.81\% \pm 0.61\%$ & $77.50\% \pm 0.13\%$ & 1.20x \tabularnewline
		R34 + 1D-SWD & $\mathbf{94.41\% \pm 0.10\%}$ & $\mathbf{77.52\% \pm 0.16\%}$ &  1.19x \tabularnewline
		\midrule
		R50  & $93.31\% \pm 0.36\%$ & $76.18\% \pm 0.76\%$ & 1.00x \tabularnewline
		R50 + 1D-SFD & $\mathbf{93.85\% \pm 0.40\%}$ & $77.38\% \pm 0.73\%$ & 1.60x \tabularnewline
		R50 + 1D-SWD & $93.67\% \pm 0.10\%$ &  $\mathbf{77.46\% \pm 0.13\%}$ & 1.25x \tabularnewline
		\midrule
		V16  & $93.27\% \pm 0.12\%$ & $72.48\% \pm 0.35\%$ & 1.00x \tabularnewline
		V16 + 1D-SFD  & $\mathbf{93.49\% \pm 0.07\%}$ & $72.27\% \pm 0.46\%$ & 1.17x \tabularnewline
		V16 + 1D-SWD  & $93.42\% \pm 0.17\%$ & $\mathbf{72.57\% \pm 0.32\%}$ & 1.08x \tabularnewline
		\bottomrule
	\end{tabular*}
\end{table}

\begin{table}[tb]
	\caption{2D Spectral Dropout Methods on CIFAR-10/100: Accuracy and Training Time Multiplier (TTM). \label{tab:cifar10-100-2D}}
	\begin{tabular*}{\linewidth}{@{\extracolsep{\fill}}lccc} 
		\toprule
		Method & CIFAR-10 Acc. & CIFAR-100 Acc. & TTM \tabularnewline
		\midrule
		R18  & $94.04\% \pm 0.08\%$ & $76.47\% \pm 0.20\%$ & 1.00x \tabularnewline
		R18 + 2D-SFD & $\mathbf{94.29\% \pm 0.19\%}$ &  $77.21\% \pm 0.31\%$ & 1.59x \tabularnewline
		R18 + 2D-SWD & $94.22\% \pm 0.16\%$ &  $\mathbf{77.22\% \pm 0.15\%}$ & 2.22x \tabularnewline
		\midrule
		R34  & $93.69\% \pm 0.30\%$  & $77.07\% \pm 0.45\%$ & 1.00x \tabularnewline
		R34 + 2D-SFD & $93.97\% \pm 0.26\%$ & $\mathbf{77.49\% \pm 0.28\%}$ & 1.30x \tabularnewline
		R34 + 2D-SWD & $\mathbf{94.03\% \pm 0.31\%}$ & $77.39\% \pm 0.36\%$ & 2.21x \tabularnewline
		\midrule
		R50  & $93.31\% \pm 0.36\%$ &  $76.18\% \pm 0.76\%$ & 1.00x \tabularnewline
		R50 + 2D-SFD & $93.59\% \pm 0.40\%$ &  $77.63\% \pm 0.55\%$ & 1.87x \tabularnewline
		R50 + 2D-SWD & $\mathbf{93.62\% \pm 0.31\%}$ & $\mathbf{77.98\% \pm 0.10\%}$ & 2.91x
		\tabularnewline
		\midrule
		V16  & $93.27\% \pm 0.12\%$ &  $72.48\% \pm 0.35\%$ & 1.00x \tabularnewline
		V16 + 2D-SFD  & $93.48\% \pm 0.12\%$ & $72.52\% \pm 0.46\%$ & 1.12x \tabularnewline
		V16 + 2D-SWD  & $\mathbf{93.50\% \pm 0.21\%}$ & $\mathbf{72.63\% \pm 0.11\%}$ & 1.88x \tabularnewline
		\bottomrule
	\end{tabular*}
\end{table}

\subsection{Image Classification on ImageNet}

We extend our evaluation to the ImageNet dataset \cite{2015_Russakovsky}, which contains 1.2 million training images, 50,000 validation images, and 150,000 test images across 1,000 categories. 
We performed our experiments on ResNet50 across four distinct seeds, training each model for 450 epochs. Following the common practice, we report the top-1 classification accuracy on the validation set.

Considering the computational efficiency of 1D-SWD and the established presence of 2D-SFD in literature, our focus is on these methods. We have focused our scope due to the computational overhead associated with 1D-SFD and 2D-SWD.
Specifically, 1D-SFD leads to a training time multiplier of 1.77x, while 2D-SWD extends this to 3.12x on ImageNet.

We also include a comparison with modern dropout techniques such as dropblock (DB) and dropout2d (D2D) \cite{2018_Ghiasi_CONF}. Comprehensive implementation details are provided in the Appendix, Section \ref{imagenet_implementation_details}. 

\subsubsection{Results}

Table \ref{tab:ImageNet} shows the effectiveness of our approach on ImageNet. For ResNet50, the application of 1D-SWD resulted in an accuracy increase of 0.60\%pt compared to the baseline. It surpasses both DB and D2D. For ResNet50, the application of 1D-SWD resulted in an accuracy increase of 0.60\%pt.

While 2D-SFD shows the highest performance gain, it requires tuning two hyperparameters, which is not needed with 1D-SWD. Additionally, the runtime overhead of 2D-SFD, noted as more than double (2.04 times) the baseline for ResNet50 on ImageNet, is significantly higher than the more manageable increase of approximately 1.28 times overhead of 1D-SWD. This difference not only highlights the computational efficiency of 1D-SWD but also its attractiveness as a simplified alternative to 2D-SFD.

\begin{table}[tb]
	\caption{Comparison of Dropout Techniques on ImageNet: Top-1 Accuracy and Training Time Multiplier (TTM). \label{tab:ImageNet}}
	\begin{tabular*}{\columnwidth}{@{\extracolsep{\fill}}lcc} 
		\toprule
		Method & Top-1 Accuracy & TTM \tabularnewline
		\midrule
		ResNet50 (Baseline) & $76.87\% \pm 0.15\%$ & 1.00x \tabularnewline
		ResNet50 + D2D & $77.27\% \pm 0.12\%$ & 1.13x \tabularnewline
		ResNet50 + DB &  $77.40\% \pm 0.11\%$ & 1.21x \tabularnewline
		ResNet50 + 1D-SWD & $\mathbf{77.47\% \pm 0.05\%}$ & 1.28x \tabularnewline
		ResNet50 + 2D-SFD & $\mathbf{77.87\% \pm 0.06\%}$ & 2.04x \tabularnewline
		\bottomrule
	\end{tabular*} 
\end{table}

\subsection{Object Detection on Pascal VOC}

We compared SWD in both 1D and 2D forms to the established 2D-SFD and its one-dimensional counterpart on object detection. We tested these methods on the PASCAL VOC dataset \cite{2010_Everingham}, which includes a diverse set of 20 object categories. We use the mmobjectdetection library \cite{2019_Chen} to train our models. 

We used the Faster R-CNN framework \cite{2015_Ren_CONF} and integrated our Spectral Wavelet Dropout approach to evaluate its impact on model performance. We selected Mean Average Precision (mAP) and Average Precision at 50\% Intersection over Union (AP50) as our main metrics. Further implementation details can be found in the Appendix, Section \ref{pascalvoc_implementation_details}.

Table \ref{tab:pascal_voc} shows the experimental results. The 2D-SWD variant achieved the highest mAP and AP50 scores, demonstrating its superiority over SFD in object detection. Notably, 1D-SWD also showed a performance improvement while having much lower training overhead compared to 1D/2D-SFD and 2D-SWD. Both 1D-SWD and 2D-SWD outperformed their 1D and 2D Spectral Fourier Dropout counterparts. This shows the effectiveness of SWD as an alternative regularization method for object detection.

\begin{table}[tb]
	\caption{Performance of Spectral Dropout Methods in Object Detection on PASCAL VOC: Mean Average Precision (mAP), Average Precision at 50\% IoU (AP50), and Training Time Multiplier (TTM). The training overhead is presented as a factor of the baseline model's computational time, where ``1.00x'' represents the baseline. \label{tab:pascal_voc}}
	\begin{tabular*}{\columnwidth}{@{\extracolsep{\fill}}lccc} 
		\toprule
		Method & mAP (\%) & AP50 (\%) & TTM \tabularnewline
		\midrule
		FRCNN (Baseline)  &   $77.61\% \pm 0.27\%$ & $77.62\% \pm 0.26\%$ & 1.00x \tabularnewline
		\midrule
		FRCNN + 1D-SFD & $77.70\% \pm 0.17\%$ & $77.68\% \pm 0.17\%$ & 3.39x \tabularnewline
		FRCNN + 1D-SWD  & $\mathbf{78.01\% \pm 0.16\%}$ & $\mathbf{78.00\% \pm 0.14\%}$ & 1.58x \tabularnewline
		\midrule
		FRCNN + 2D-SFD  & $77.86\% \pm 0.11\%$ & $77.86\% \pm 0.10\%$ & 4.07x \tabularnewline
		FRCNN + 2D-SWD & $\mathbf{78.17\% \pm 0.18\%}$ & $\mathbf{78.16\% \pm 0.19\%}$ & 3.51x \tabularnewline
		\bottomrule
	\end{tabular*}
\end{table}

\section{Analysis and Discussion}

\subsection{Hyperparameters}

For detailed outcomes of the hyperparameter optimization on CIFAR-10 and CIFAR-100, refer to Tables \ref{tab:hyperparameters_1D} and \ref{tab:hyperparameters_2D} in Appendix \ref{hyperparameters_paper}. Results for ImageNet are presented in Table \ref{tab:hyperparameters_imageNet}. For SWD it is preferred to use a small dropout rate, i.e.\ $p=\{0.1, 0.2\}$. Increasing the dropout rate resulted in a decrease in the overall performance. In contrast to SWD, SFD needs different dropout and pruning rates for each setting.

\subsection{Frequency Dropout - 1D vs 2D}

The decision to use one-dimensional (1D) or two-dimensional (2D) Spectral Dropout, which includes both Wavelet and Fourier approaches, depends on their decomposition strategy. 1D methods target horizontal features, which are effectively for tasks with dominant horizontal patterns. Conversely, 2D techniques analyze both vertical and horizontal dimensions, therefore enhancing feature extraction at the cost of increased computational demand. Our findings show a trade-off between computational efficiency and performance enhancement that is influenced by the complexity of the task. For less demanding tasks, where the trade-off is minimal, we recommend the use of 1D Spectral Dropout due to its efficiency. In more challenging the choice between 2D Spectral Dropout for optimal performance and 1D Spectral Dropout as a more computationally efficient regularization method must be considered carefully.

\subsection{Computational Complexity Analysis} \label{comp_complexity}

We compare the computational complexities and runtime performances of Spectral Wavelet Dropout and Spectral Fourier Dropout. Let $n$ be the number of rows/columns of a square matrix. Theoretically, 2D-SFD and 1D-SFD have complexities of $\mathcal{O}(n^2\log n)$ for frequency transformations, indicating higher computational demands as the feature map sizes increase \cite{1997_Guo_CONF}. Conversely, SWD, in both 1D and 2D forms, operates at a reduced complexity of $\mathcal{O}(n^2)$. This improves efficiency due to a direct correlation with the pixel count.

Details on the runtime overheads can be found in the ``Train Time Multiplier'' column across Tables \ref{tab:cifar10-100-1D}, \ref{tab:cifar10-100-2D}, \ref{tab:ImageNet}, and \ref{tab:pascal_voc}. The runtime overhead of the 2D-SWD variant exceed that of 2D Spectral Fourier Dropout, likely due to less optimized wavelet transform algorithms or suboptimal GPU utilization. Notably, the Fourier Transform, known for its higher theoretical computational demands, showed improved efficiency on smaller datasets. This is attributed to the utilization of optimized GPU-accelerated libraries. 

Our findings show that 1D-SWD consistently has a lower runtime overheads than both 1D and 2D-SFD on image classification and object detection task. Note that the runtime overhead by SWD is limited to the training phase and does not affect the inference time.

\subsection{Dropping Fourier Coefficients vs Frequency Bands}

A main difference between SWD and SFD is the type of the dropped
values. Specifically, SWD targets frequency bands of a wavelet transformed feature map, while SFD drops random Fourier coefficients in a Fourier-transformed feature map.

An advantage of dropping specific frequency bands is the possibility to control
the level of regularization. By choosing specific frequency bands of the wavelet
decomposition and adjusting the dropout rate, one can fine-tune the
regularization strength based on the characteristics of the task and the desired
model complexity. 

In contrast to the Fourier transform, the Wavelet transform allows a
multi-resolution analysis. Dropping the frequency bands forces the model to
use information at different frequency scales. 

Moreover, not all Fourier coefficients contribute equally to the information
content of the image. Randomly dropping coefficients may unintentionally remove
important frequency components, leading to a potential loss of crucial
information. Furthermore, such random elimination could also have no
effect at all, if the Fourier coefficients do not contribute to the information
content of the image.

\section{Conclusion and Future Work}

In this work, we introduced Spectral Wavelet Dropout, a novel regularization technique designed to improve the generalization capabilities of CNNs by randomly dropping frequency bands in wavelet-transformed feature maps. Spectral Wavelet Dropout is presented in two variants: 1D-SWD and 2D-SWD, which use different wavelet transformations.

Empirical evaluations show that Spectral Wavelet Dropout matches and in some cases surpasses Spectral Fourier Dropout on datasets such as CIFAR-10, CIFAR-100 and ImageNet. Notably, the 1D variant of SWD achieves this competitive performance with a significant reduction in computational complexity, distinguishing it from both its 2D counterpart and SFD alternatives. In addition, SWD simplifies optimization by using just one hyperparameter, unlike the two required by SFD. For object detection tasks on the Pascal VOC dataset, both the 1D and 2D variants of Spectral Wavelet Dropout perform better than their Spectral Fourier Dropout counterparts while requiring less computation. Moreover, SWD requires only one hyperparameter instead of two in SFD.

While our current investigation has shown promising results, the full potential of SWD remains to be fully unlocked. Future research directions include the investigation of different wavelet functions, exploration of adaptive wavelets, and deeper analysis of decomposition levels to further refine the efficacy of SWD. Additionally, optimizing the handling of border distortions and exploring selective regularization based on frequency band energy distribution could provide avenues for enhancing the performance and applicability of SWD.

\bibliography{Wavelet_Dropout}
\bibliographystyle{IEEEtran}

\newpage

\section{Appendix}

\FloatBarrier

\subsection{Visualizing 1D Spectral Wavelet Dropout}

\begin{figure*}[tb]
	\includegraphics[width=\linewidth]{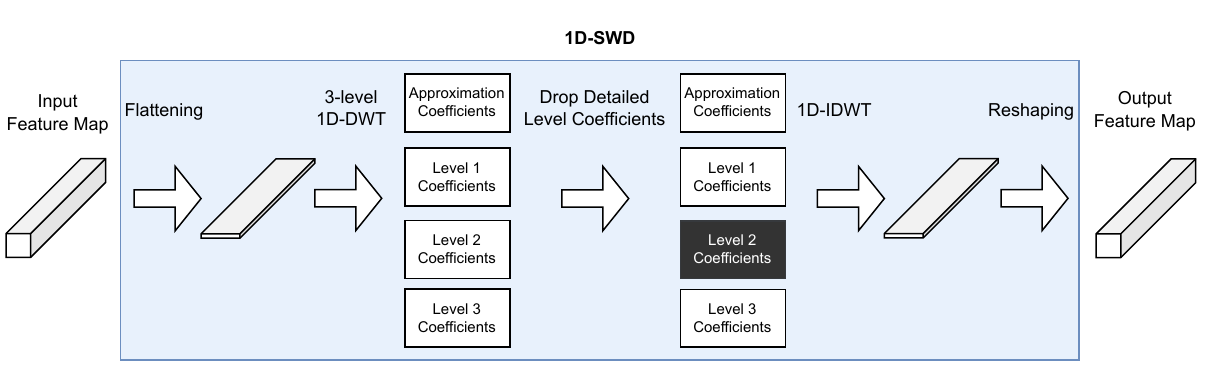}
	\caption{Schematic of the 1D-SWD applied to a feature map \( X \in \mathbb{R}^{C \times H \times W} \). The process begins by flattening \( X \) across channels, resulting in \( \hat{X} \in \mathbb{R}^{C \times H \cdot W} \). A three-level 1D-DWT is then applied to \( \hat{X} \), yielding approximation and detailed coefficients across three levels. For regularization, detailed coefficients are dropped using a dropout probability \( p \), with Level 2 Coefficients being dropped in the example. The regularized feature map is transformed back using an inverse 1D-DWT.}
	\label{fig:SWD_Layer_app}
\end{figure*}

Figure~\ref{fig:SWD_Layer_app} illustrates the 1D Spectral Wavelet Dropout applied to a feature map \( X \in \mathbb{R}^{C \times H \times W} \). The process begins by flattening the feature map along each channel, resulting in \( \hat{X} \in \mathbb{R}^{C \times H \cdot W} \). Then a three-level one-dimensional discrete wavelet decomposition (1D-DWT) is performed on \( \hat{X} \), resulting in approximation and detailed frequency coefficients across three levels. They are also referred as frequency bands. In the next phase, 1D-SWD randomly drops detailed wavelet bands using a dropout probability $p$. This introduces regularization. This regularization is shown in Figure~\ref{fig:SWD_Layer_app} by the dropping of the Level 2 coefficients. The feature map is transformed back into the spatial domain using an inverse 1D-DWT. Finally, it is reshaped for input into the subsequent network layer.

\subsection{Implementing the 2D Spectral Wavelet Dropout Algorithm} \label{algorithm_2d_dwt}

At the core of the 2D-SWD, as detailed in Algorithm \ref{algorith_2D_SWD}, is the application of the 2D Discrete Wavelet Transform (2D-DWT). This transformation process decomposes each channel of the input feature map into four distinct sub-bands: $LL$ (low-low), $LH$ (low-high), $HL$ (high-low), and $HH$ (high-high). These represent the approximation, horizontal detail, vertical detail, and diagonal detail components.

Following the decomposition, a dropout mask is sampled using a dropout probability $p$. This mask is applied to the detail coefficients ($LH$, $HL$, and $HH$). This drops a portion of the wavelet coefficients to introduce regularization. The approximation coefficients ($LL$) are preserved to maintain the overall structure of the image. Scaling the rest of the coefficients by a factor of $(1-p)^{-1}$ compensates for the dropout's reduction of signal energy.

The final step involves transforming the feature map back into the spatial domain using the inverse 2D-DWT.

\begin{algorithm}[!ht] 
	\DontPrintSemicolon
	\Parameter{\text{dropout probability = }$p \in (0,1)$}
	\KwInput{$X \in \mathbb{R}^{B \times C \times H \times W}$}
	\KwOutput{$\hat{X} \in \mathbb{R}^{B \times C \times H \times W}$}
	
	\tcp{Save size of $X$}
	size = $X$.size()
	
	\tcp{Compute the 1-level 2D-DWT for each channel using the db3-wavelet}
	$LL, LH, HL, HH = \text{2D-DWT}(X, J=1, \text{wave='db3'})$
	
	\tcp{Create mask for dropout}
	
	$\text{dropout\_mask} = \text{Bernoulli}(3, \text{prob}=1-p)$
	
	\tcp{Wavelet Dropout}
	
	$LH = LH \cdot \text{dropout\_mask}[0]/(1-p)$
	
	$HL = HL \cdot \text{dropout\_mask}[1]/(1-p)$
	
	$HH = HH \cdot \text{dropout\_mask}[2]/(1-p)$
	
	\tcp{Compute inverse Wavelet transformation}
	
	$\hat{X} =  \text{2D-IDWT}([LL, LH, HL, HH], \text{wave='db3'})$
	
	\caption{Spectral Wavelet Dropout - 2D-SWD} \label{algorith_2D_SWD}
\end{algorithm}

\subsection{In-Depth Analysis of the Discrete Wavelet Transform in SWD}

\begin{figure}
	\centering
	\includegraphics[width=\columnwidth]{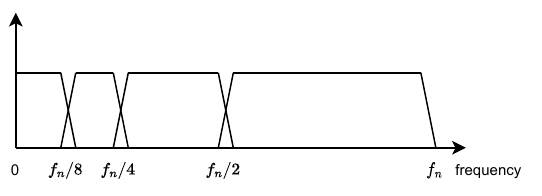}
	\caption{This figure shows the Discrete Wavelet Transform dividing a signal into distinct frequency bands. Such multi-resolution analysis is important for our Spectral Wavelet Dropout method, as it allows randomly dropping of information across different scales.}
	\label{fig:DWT_Freq}
\end{figure}

Figure \ref{fig:DWT_Freq} illustrates the frequency domain representation of the Discrete Wavelet Transform. It demonstrates the DWT's ability to partition a signal into multiple frequency components.

\subsection{Detailed CIFAR-10/100 Training Procedures} \label{cifar_implementation_details}

Our CIFAR experiments used PyTorch 1.10.1 \cite{2019_Paszke_CONF} on an Nvidia GeForce 1080Ti GPU. We used the same training setup for all models, with a batch size of 128 and the SGD optimizer with a momentum of 0.9. The initial learning rate was set to 0.1, with adjustments based on the dataset: for CIFAR-10, it was reduced by a factor of 0.1 at epochs 90 and 136, along with a weight decay of $1e-4$ \cite{2016_He_CONF}; for CIFAR-100, it was reduced by 0.2 at epochs 60, 120, and 160, using a weight decay of $5e-4$. These parameter settings match those recommended in \cite{2016_He_CONF}, without further hyperparameter optimization. Data augmentation techniques included random cropping (size 32 with padding 4), random horizontal flipping, and normalization, as outlined in \cite{2019_Shorten}, with normalization being the only preprocessing step for the test set. Network initialization followed the Kaiming uniform approach \cite{2015_He_CONF}, with VGG's linear layers similarly initialized and its convolutional layers set to a Gaussian distribution ($\mu=0$, $\sigma=\sqrt{2/n}$).

\subsection{Detailed ImageNet Training Procedures} \label{imagenet_implementation_details}

In our experiments, we used PyTorch 1.10.1 \cite{2019_Paszke_CONF} with four Nvidia GeForce 1080Ti GPUs. All models were trained with the SGD optimizer, set with a momentum of $0.9$ and an initial learning rate of $0.1$, which was reduced by a factor of $0.1$ at epochs $75$, $150$, $225$, $300$, and $375$. A weight decay factor of $1e-4$ was applied, and the batch size was standardized at $128$.

For data augmentation, the training dataset was randomly cropped to $224\times224$, randomly flipped horizontally, and normalized. The validation dataset was resized to $256\times256$, center cropped to $224\times224$, and normalized. Our setup matches the configuration specified in the PyTorch ImageNet training example\footnote{\url{https://github.com/pytorch/examples/blob/main/imagenet}}.

\subsection{Implementation Details on PASCAL VOC using Faster R-CNN}
\label{pascalvoc_implementation_details}

In our object detection experiments on the PASCAL VOC dataset, we use the mmobjectdetection library \cite{2019_Chen} for setting up and training our models. This section outlines the key implementation details to ensure our results can be reproduced.

\paragraph{Model Configuration} We set up our Faster R-CNN \cite{2015_Ren_CONF} with a ResNet-50 backbone and a Feature Pyramid Network (FPN) for multi-scale feature extraction.

\paragraph{Data Preprocessing and Augmentation} Our preprocessing pipeline includes several steps to prepare images for object detection. First, images are loaded and annotations are retrieved. Then, images are resized to $(1000, 600)$, keeping their original aspect ratio. To augment the data and introduce variability, we randomly flip images horizontally with a 50\% probability. This augmentation is applied uniformly across the training dataset to enhance model robustness and generalization. For validation, images are resized in the same way but without random flips to keep evaluation consistent.

\paragraph{Training Configuration} The model is trained on the combined trainval sets of VOC2007 and VOC2012 and evaluated on the VOC2007 validation set. We use a batch size of 2. The SGD optimizer is used with momentum and weight decay, and the learning rate is adjusted according to a predefined schedule. We use the mean Average Precision (mAP) metric, calculated with the `11points` interpolation method, as the evaluation metric.

\paragraph{Spectral Dropout Integration} All Spectral Dropout methods were applied within the third block of the ResNet50 backbone.

\subsection{Frequency Band Impact Analysis with 1D-SWD}

We analyzed the effects of explicitly dropping different frequency components: the approximation coefficients, and level 1, 2, and 3 detail coefficients, both individually and in combinations. Our results are shown in Tables \ref{tab:swd-cifar10_HH}, \ref{tab:swd-cifar100_HH}, and \ref{tab:swd-imagenet_HH}. The naming convention of model configurations, such as ``ResNet50 + 1D-SWD34'', indicates where 1D-SWD is applied within the ResNet50 architecture, showing the specific stages where frequency band manipulation happens. 

They indicate a significant performance improvement when only level 3 detail coefficients are dropped across CIFAR-10/100 and ImageNet datasets. Our analysis shows that dropping approximation coefficients, which capture the low-frequency, smoothed parts of the signal, is not effective. These coefficients contain important information about the signal's overall structure, whereas detail coefficients correspond to high-frequency variations and potential noise.

\begin{table}[tb]
	\caption{Performance on CIFAR-10 with 1D-SWD Targeting Level 3 Detail Coefficients. \label{tab:swd-cifar10_HH}}
	\begin{tabular*}{\columnwidth}{@{\extracolsep{\fill}}lc} 
		\toprule
		\textbf{Method} & \textbf{Accuracy} \tabularnewline
		\midrule
		ResNet18 + 1D-SWD$3$ & $94.05\% \pm 0.21\%$ \tabularnewline
		\midrule
		ResNet34 + 1D-SWD$3$ & $94.26\% \pm 0.23\%$ \tabularnewline
		\midrule
		ResNet50 + 1D-SWD$34$ & $94.00\% \pm 0.18\%$ \tabularnewline
		\bottomrule
	\end{tabular*}
\end{table}

\begin{table}[tb]
	\caption{Performance on CIFAR-100 with 1D-SWD Targeting Level 3 Detail Coefficients. \label{tab:swd-cifar100_HH}}
	\begin{tabular*}{\columnwidth}{@{\extracolsep{\fill}}lc} 
		\toprule
		\textbf{Method} & \textbf{Accuracy} \tabularnewline
		\midrule
		ResNet18 + 1D-SWD$4$ & $77.11\% \pm 0.21\%$ \tabularnewline
		\midrule
		ResNet34 + 1D-SWD$4$ & $77.32\% \pm 0.41\%$ \tabularnewline
		\midrule
		ResNet50 + 1D-SWD$4$ & $77.42\% \pm 0.47\%$ \tabularnewline
		\bottomrule
	\end{tabular*}
\end{table}

\begin{table}[tb]
	\caption{Performance on ImageNet with 1D-SWD Targeting Level 3 Detail Coefficients. \label{tab:swd-imagenet_HH}}
	\begin{tabular*}{\columnwidth}{@{\extracolsep{\fill}}lc} 
		\toprule
		\textbf{Method} & \textbf{Top-1 Accuracy} \tabularnewline
		\midrule
		ResNet50 + 1D-SWD$4$ & $77.51\% \pm 0.11\%$ \tabularnewline
		\bottomrule
	\end{tabular*}
\end{table}

\subsection{Hyperparameters} \label{hyperparameters_paper}

\subsubsection{Method Parameters}

The hyperparameters for our CIFAR and ImageNet experiments are shown in Tables \ref{tab:hyperparameters_1D}, \ref{tab:hyperparameters_2D}, and \ref{tab:hyperparameters_imageNet}. We found that lower dropout rates, specifically $p=\{0.1, 0.2\}$, work best for both 1D and 2D Spectral Wavelet Dropout. Higher dropout rates led to lower performance.

In contrast, Spectral Fourier Dropout in both 1D and 2D forms required different dropout and pruning rates, showing the need for specific hyperparameter tuning for each setting.

\begin{figure}[h]
	\includegraphics[width=\columnwidth]{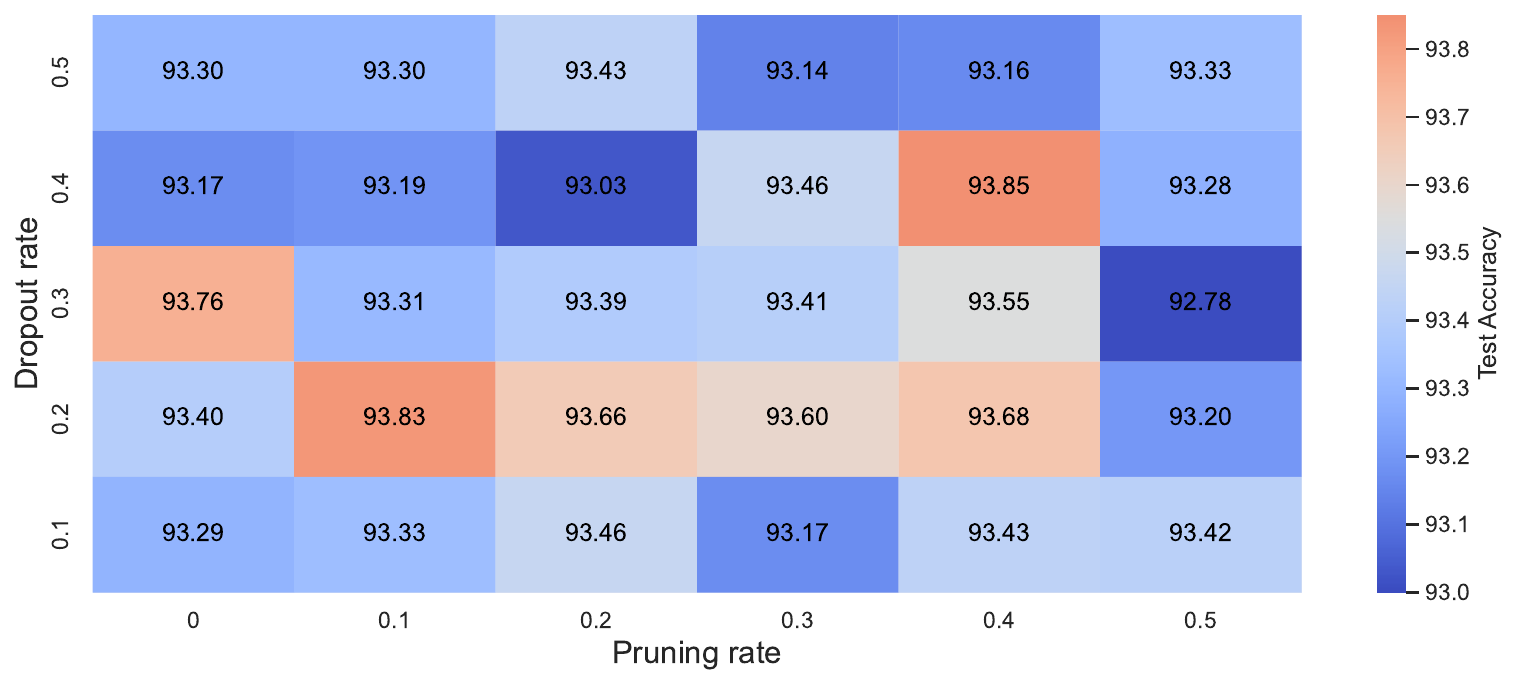}
	\caption{Hyperparameter search results for ResNet50 with 2D-SFD on CIFAR-10, illustrating the impact of various configurations on model performance.}
	\label{fig:hyp_search_resNet50_cifar10}
\end{figure}

\begin{figure}[h]
	\includegraphics[width=\columnwidth]{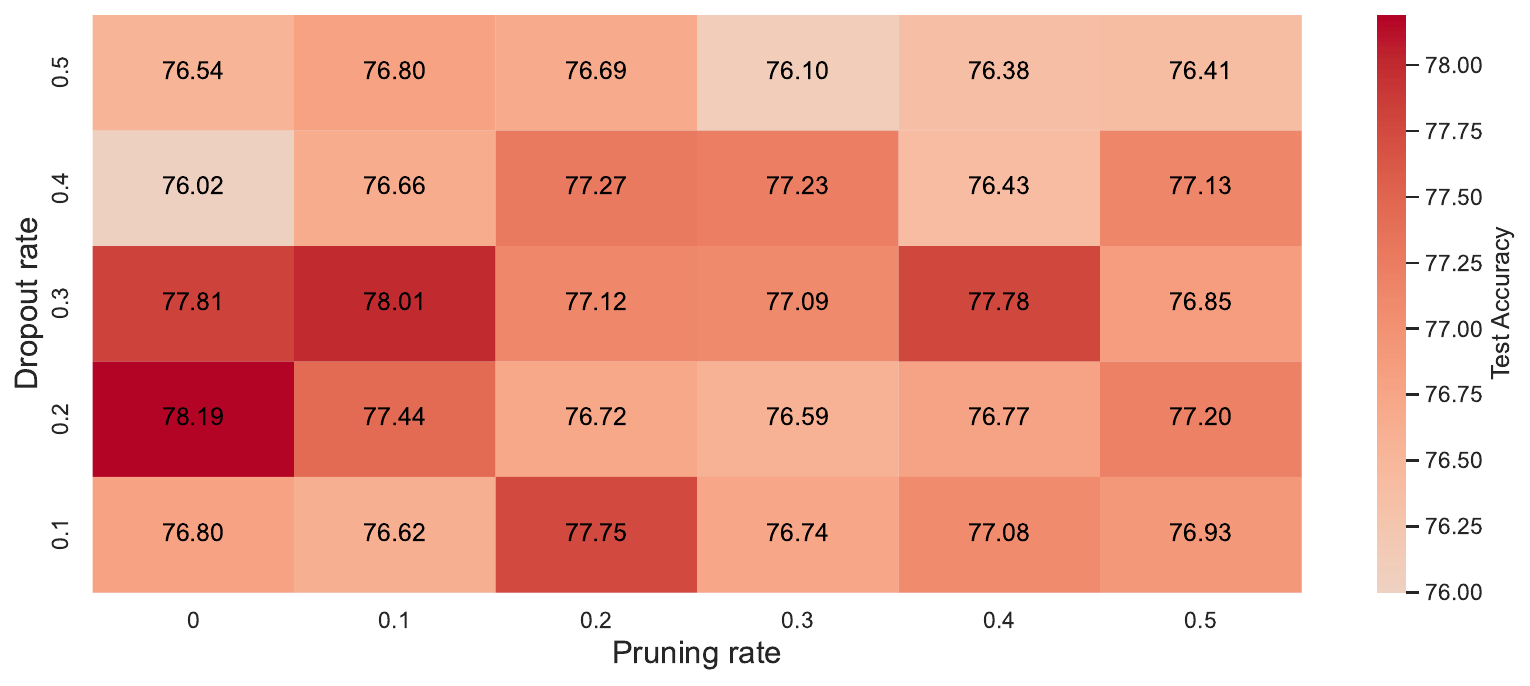}
	\caption{Hyperparameter search results for ResNet50 with 2D-SFD on CIFAR-100, demonstrating the performance variability across different hyperparameter settings.}
	\label{fig:hyp_search_resnet50_cifar100}
\end{figure}

Figures \ref{fig:hyp_search_resNet50_cifar10} and \ref{fig:hyp_search_resnet50_cifar100} show how ResNet50 performance on CIFAR-10 and CIFAR-100 varies with different hyperparameters for 2D-SFD. This underscores the importance of a comprehensive hyperparameter search for SFD.

\begin{table}[tb]
	\caption{Hyperparameters for CIFAR-10 and CIFAR-100 experiments with 1D spectral methods. \label{tab:hyperparameters_1D}}
	\begin{tabular*}{\linewidth}{@{\extracolsep{\fill}}lcc} 
		\toprule
		\textbf{Model} & \textbf{CIFAR-$10$} & \textbf{CIFAR-$100$} \tabularnewline
		\midrule
		ResNet18 + 1D-SWD & $p=0.1$ & $p=0.1$ \tabularnewline
		ResNet18 + 1D-SFD & $p=0.1,\eta=0.1$ & $p=0.2,\eta=0.2$ \tabularnewline
		\midrule
		ResNet34 + 1D-SWD & $p=0.2$ & $p=0.1$\tabularnewline
		ResNet34 + 1D-SFD & $p=0.3,\eta=0.5$ & $p=0.1,\eta=0.0$ \tabularnewline
		\midrule
		ResNet50 + 1D-SWD & $p=0.2$ & $p=0.1$ \tabularnewline
		ResNet50 + 1D-SFD & $p=0.2,\eta=0.1$ & $p=0.3, \eta=0.1$ \tabularnewline
		\midrule 
		VGG16 + 1D-SWD & $p=0.1$ & $p=0.2$ \tabularnewline
		VGG16 + 1D-SFD & $p=0.3, \eta=0.4$ & $p=0.4, \eta=0.1$ \tabularnewline
		\bottomrule
	\end{tabular*}
\end{table}

\begin{table}[h]
	\caption{Hyperparameters for CIFAR-10 and CIFAR-100 experiments with 2D spectral methods. \label{tab:hyperparameters_2D}}
	\begin{tabular*}{\linewidth}{@{\extracolsep{\fill}}lcc} 
		\toprule
		\textbf{Model} & \textbf{CIFAR-$10$} & \textbf{CIFAR-$100$} \tabularnewline
		\midrule
		ResNet18 + 2D-SWD & $p=0.1$ & $p=0.2$ \tabularnewline
		ResNet18 + 2D-SFD & $p=0.1,\eta=0.2$ & $p=0.1,\eta=0.1$ \tabularnewline
		\midrule
		ResNet34 + 2D-SWD & $p=0.2$ & $p=0.2$\tabularnewline
		ResNet34 + 2D-SFD & $p=0.4,\eta=0.3$ & $p=0.2,\eta=0.0$ \tabularnewline
		\midrule
		ResNet50 + 2D-SWD & $p=0.2$ & $p=0.1$ \tabularnewline
		ResNet50 + 2D-SFD & $p=0.4,\eta=0.4$ & $p=0.2, \eta=0.0$ \tabularnewline
		\midrule 
		VGG16 + 2D-SWD & $p=0.5$ & $p=0.2$ \tabularnewline
		VGG16 + 2D-SFD & $p=0.1, \eta=0.4$ & $p=0.4, \eta=0.5$ \tabularnewline
		\bottomrule
	\end{tabular*}
\end{table}

\begin{table}[h]
	\caption{Hyperparameters of experiments on ImageNet. \label{tab:hyperparameters_imageNet}}
	\begin{tabular*}{\linewidth}{@{\extracolsep{\fill}}lc} 
		\toprule
		\textbf{Model} & \textbf{ImageNet} \tabularnewline
		\midrule
		ResNet50 + 1D-SWD & $p=0.2$ \tabularnewline
		ResNet50 + 2D-SFD & $p=0.2,\eta=0.1$ \tabularnewline
		\bottomrule
	\end{tabular*}
\end{table}

\subsubsection{Positional Analysis} \label{positional_analysis_appendix}

We did an ablation study to investigate the effect of the position of 1D-SWD
within the ResNets. We did the same ablation study for 1D-SFD, 2D-SWD and 2D-SFD. 
Similar to other dropping methods like DropBlock \cite{2018_Ghiasi_CONF} it is
preferred to insert the frequency dropout methods in deeper layers, e.g.\ for
ResNets in the modules 3 or 4. The Tables \ref{tab:cifar10-100-1D_pos} and
\ref{tab:cifar10-100-2D_pos} show in which module the frequency dropout methods
are inserted in our experiments.

The regularization methods can be too harsh
when applied to higher resolution feature maps in earlier modules. Inserting the
frequency dropout methods in the fourth module generally led to the best
results. For smaller networks on easier tasks (e.g. ResNet18 on CIFAR10) it is
preferred to insert them in the third module. This may be attributed due to the
shallower architecture, which might need more regularization in the third module
to prevent overfitting to noise or irrelevant details. In deeper ResNet
variants, the increased model capacity enables them to capture finer details
and patterns in the fourth module, making 1D-SWD more effective at this stage.

We also analyzed the effect of the position of 1D-SWD within the basic or bottleneck
buildings blocks. We investigated inserting SWD before and after the convolution
layers within the buildings blocks. Both ways increase the performance. 
However, the optimal position depends on the resolution of the feature maps.

For high-resolution images like ImageNet, 1D-SWD should be inserted before each convolution layer, following the sequence: 1D-SWD - Conv - BN - ReLU, including in the residual connection (1D-SWD - Conv - BN). This architecture forces the network to focus on various frequency bands. 

For low-resolution images like CIFAR, it is preferred to place 1D-SWD after the convolution layers (Conv - 1D-SWD - BN - ReLU), both in the main and residual block. This allows the network to first extract features through convolution, which are then manipulated using 1D-SWD.

For both low- and high-resolution images, the 2D-SWD is strategically positioned before each convolutional layer to optimize performance.

\begin{table}[h]
	\caption{Position of Spectral Dropout on CIFAR-$10$/$100$ - 1D. \label{tab:cifar10-100-1D_pos}}
	\begin{tabular*}{\linewidth}{@{\extracolsep{\fill}}lcc} 
		\toprule
		Model & CIFAR-$10$ & CIFAR-$100$ \tabularnewline
		\midrule
		R18 + 1D-SFD & 4 & 4 \tabularnewline
		R18 + 1D-SWD & 3 & 4 \tabularnewline
		\midrule
		R34 + 1D-SFD & 4 & 4 \tabularnewline
		R34 + 1D-SWD & 3 & 4 \tabularnewline
		\midrule
		R50 + 1D-SFD & 4 & 4 \tabularnewline
		R50 + 1D-SWD & 4 & 4 \tabularnewline
		\midrule
		V16 + 1D-SWD & 4 & 4 \tabularnewline
		\bottomrule
	\end{tabular*}
\end{table}

\begin{table}[h]
	\caption{Position of Spectral Dropout on CIFAR-$10$/$100$ - 2D. \label{tab:cifar10-100-2D_pos}}
	\begin{tabular*}{\linewidth}{@{\extracolsep{\fill}}lcc} 
		\toprule
		Model & CIFAR-$10$ & CIFAR-$100$ \tabularnewline
		\midrule
		R18 + 2D-SFD & 4 & 4 \tabularnewline
		R18 + 2D-SWD & 3 & 4\tabularnewline
		\midrule
		R34 + 2D-SFD & 4 & 4 \tabularnewline
		R34 + 2D-SWD & 4 & 4 \tabularnewline
		\midrule
		R50 + 2D-SFD & 4 & 4 \tabularnewline
		R50 + 2D-SWD & 4 & 4
		\tabularnewline
		\midrule
		V16 + 2D-SWD & 4 & 4 \tabularnewline
		\bottomrule
	\end{tabular*}
\end{table}

\end{document}